\documentclass[sigconf, nonacm, screen]{acmart}
\usepackage[utf8]{inputenc}
\usepackage{xspace}
\usepackage{xparse}
\usepackage{nicefrac}
\usepackage{soul}
\usepackage[capitalise]{cleveref}
\usepackage{amsmath}
\usepackage{amsfonts}
\usepackage{bm}
\usepackage{siunitx}
\usepackage{lipsum}
\usepackage{enumitem}
\usepackage[labelformat=simple]{subcaption}
\usepackage{dblfloatfix}      
\usepackage{tabu}
\usepackage{tabulary}
\usepackage{multirow}
\usepackage{multicol}
\usepackage{booktabs}
\usepackage{adjustbox}
\usepackage[flushleft]{threeparttable}
\usepackage{algorithm}
\usepackage{algpseudocodex}
\usepackage{varwidth}
\usepackage{environ}
\usepackage[many]{tcolorbox}
\usepackage{makecell,tabularx}
\newcolumntype{L}{>{\raggedright\arraybackslash}X}
\newlength{\bubblesep}
\newlength{\bubblewidth}
\AtBeginDocument{\setlength{\bubblewidth}{.75\textwidth}}
\definecolor{bubblegreen}{RGB}{227,240,210}
\definecolor{bubblegray}{RGB}{241,240,240}

\newcommand{\bubble}[4]{%
  \tcbox[
    boxsep=0.5mm,
    top=1.2mm,
    bottom=1.2mm,
    left=2ex,
    right=2ex,
    on line,
    arc=2.4mm,
    colback=#1,
    colframe=#1,
    tcbox width=auto limited,
    width=0.95\columnwidth,
    #2,
  ]{\color{#3}#4}%
}

\ExplSyntaxOn
\seq_new:N \l__ooker_bubbles_seq
\tl_new:N \l__ooker_bubbles_first_tl
\tl_new:N \l__ooker_bubbles_last_tl

\NewEnviron{rightbubbles}
 {
  \begin{flushright}
  \itshape\small
  \seq_set_split:NnV \l__ooker_bubbles_seq { \par } \BODY
  \int_compare:nTF { \seq_count:N \l__ooker_bubbles_seq < 2 }
   {
    \bubble{bubblegreen}{rounded~corners}{black}{\BODY}\par
   }
   {
    \seq_pop_left:NN \l__ooker_bubbles_seq \l__ooker_bubbles_first_tl
    \seq_pop_right:NN \l__ooker_bubbles_seq \l__ooker_bubbles_last_tl
    \bubble{bubblegreen}{sharp~corners=southeast}{white}{\l__ooker_bubbles_first_tl}
    \par\nointerlineskip
    \addvspace{\bubblesep}
    \seq_map_inline:Nn \l__ooker_bubbles_seq
     {
      \bubble{bubblegreen}{sharp~corners=east}{white}{##1}
      \par\nointerlineskip
      \addvspace{\bubblesep}
     }
    \bubble{bubblegreen}{sharp~corners=northeast}{white}{\l__ooker_bubbles_last_tl}
    \par
   }
   \end{flushright}
 }
\NewEnviron{leftbubbles}
 {
  \begin{flushleft}
  \itshape\small
  \seq_set_split:NnV \l__ooker_bubbles_seq { \par } \BODY
  \int_compare:nTF { \seq_count:N \l__ooker_bubbles_seq < 2 }
   {
    \bubble{bubblegray}{rounded~corners}{black}{\BODY}\par
   }
   {
    \seq_pop_left:NN \l__ooker_bubbles_seq \l__ooker_bubbles_first_tl
    \seq_pop_right:NN \l__ooker_bubbles_seq \l__ooker_bubbles_last_tl
    \bubble{bubblegray}{sharp~corners=southwest}{black}{\l__ooker_bubbles_first_tl}
    \par\nointerlineskip
    \addvspace{\bubblesep}
    \seq_map_inline:Nn \l__ooker_bubbles_seq
     {
      \bubble{bubblegray}{sharp~corners=west}{black}{##1}
      \par\nointerlineskip
      \addvspace{\bubblesep}
     }
    \bubble{bubblegray}{sharp~corners=northwest}{black}{\l__ooker_bubbles_last_tl}\par
   }
  \end{flushleft}
 }
\ExplSyntaxOff
\newcommand{\hlc}[2][yellow]{{%
    \colorlet{foo}{#1}%
    \sethlcolor{foo}\hl{#2}}%
}

\crefname{figure}{Figure}{Figures}
\graphicspath{{figs/}}
\DeclareGraphicsExtensions{.pdf,.eps,.png,.jpg,.jpeg}

\NewDocumentCommand{\leo}{mg}{\IfNoValueTF{#2}
{\textcolor{teal}{#1}}
{\textcolor{teal}{$\blacktriangleright$}#1 \textcolor{teal}{$\triangleright$ #2$\blacktriangleleft$}}}
\newcommand{\name}{\textsc{SEQ-GPT}}

\newcommand\vldbdoi{XX.XX/XXX.XX}
\newcommand\vldbpages{XXX-XXX}
\newcommand\vldbvolume{14}
\newcommand\vldbissue{1}
\newcommand\vldbyear{2020}
\newcommand\vldbauthors{\authors}
\newcommand\vldbtitle{\shorttitle}
\newcommand\vldbavailabilityurl{URL_TO_YOUR_ARTIFACTS}
\newcommand\vldbpagestyle{plain}

\begin{document}
\title{\name{}: LLM‐assisted Spatial Query via Example}

\author{Ivan Khai Ze Lim$^{\ast}$}
\myauthornote{Both authors contributed equally to this research.}
\affiliation{%
  \institution{Nanyang Technological University}
}
\email{ilim017@e.ntu.edu.sg}

\author{Ningyi Liao$^{\ast}$}
\affiliation{%
  \institution{Nanyang Technological University}
}
\email{liao0090@e.ntu.edu.sg}

\author{Yiming Yang}
\affiliation{%
  \institution{Nanyang Technological University}
}
\email{yang0700@e.ntu.edu.sg}

\author{Gerald Wei Yong Yip}
\affiliation{%
  \institution{Nanyang Technological University}
}
\email{gyip002@e.ntu.edu.sg}

\author{Siqiang Luo}
\affiliation{%
  \institution{Nanyang Technological University}
}
\email{siqiang.luo@ntu.edu.sg}

\begin{abstract}
Contemporary spatial services such as online maps predominantly rely on user queries for location searches. However, the user experience is limited when performing complex tasks, such as searching for a group of locations simultaneously. In this study, we examine the extended scenario known as Spatial Exemplar Query (SEQ), where multiple relevant locations are jointly searched based on user-specified examples. We introduce \name{}, a spatial query system powered by Large Language Models (LLMs) towards more versatile SEQ search using natural language.
The language capabilities of LLMs enable unique interactive operations in the SEQ process, including asking users to clarify query details and dynamically adjusting the search based on user feedback. We also propose a tailored LLM adaptation pipeline that aligns natural language with structured spatial data and queries through dialogue synthesis and multi-model cooperation. \name{} offers an end-to-end demonstration for broadening spatial search with realistic data and application scenarios.
\end{abstract}

\maketitle

\begin{figure}[t]
    \centering
    \vspace*{6pt}
    \begin{subfigure}[b]{\columnwidth}
        \centering
        \includegraphics[width=0.95\textwidth]{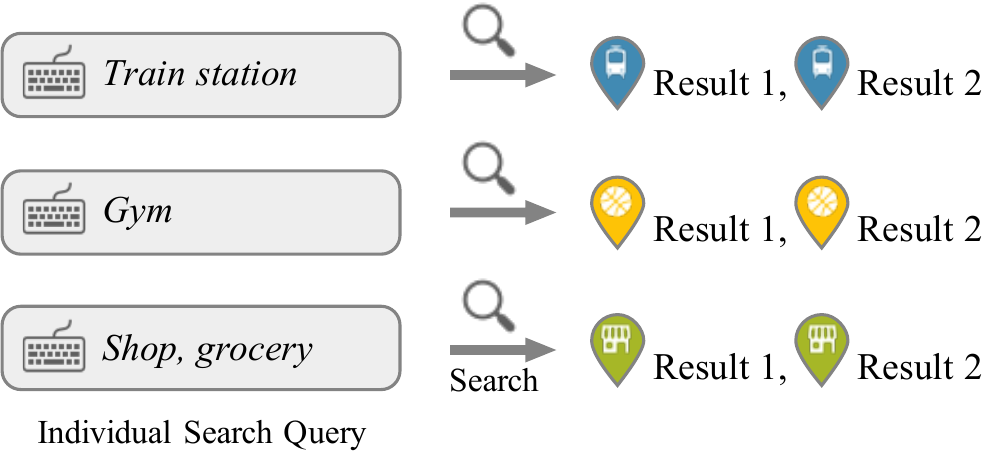}
        \caption{Conventional search for multiple locations.}
        \label{fig:example_old}
    \end{subfigure}
    \\[6pt]
    \begin{subfigure}[b]{\columnwidth}
        \centering
        \includegraphics[width=0.92\textwidth]{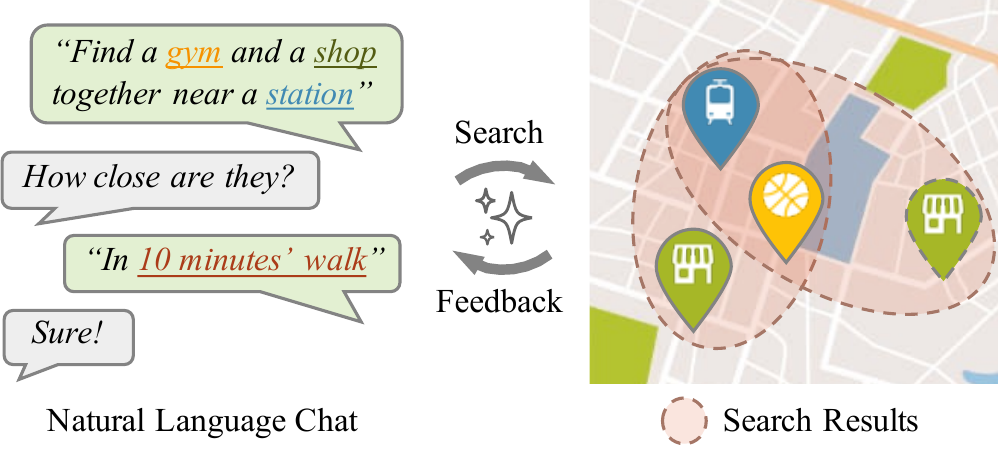}
        \caption{Spatial Exemplar Query powered by LLM}
        \label{fig:example_seq}
    \end{subfigure}
    \caption{Illustration of (a) a sequence of independent conventional spatial searches for multiple queried locations; and (b) SEQ as conversation with LLM using natural language. Compared to conventional search criteria in plain text, natural language queries allow the system to justify query details through real-time interaction with the user.}
    \label{fig:example}
    \vspace{-16pt}
\end{figure}

\section{Introduction}
\label{sec:introduction}

Spatial services have become indispensable in today’s world, powering a wide range of applications such as taxi-hailing~\cite{xu2014taxi,hwang2015effective,luo2018toain}, shortest-path routing~\cite{wu2012shortest}, POI recommendation~\cite{zhao2020go,yin2017spatial,qin2023diffusion}, and spatial example queries~\cite{luo2017seq,zhang2022example}. This paper presents a new system that integrates large language models (LLMs) into the framework of spatial example queries (SEQ).

\noindentparagraph{Reimagining Spatial Query via Example.}
Spatial search is widely used in daily scenarios such as online map services. Conventionally, it operates by filtering locations based on isolated criteria based on user-defined constraints, and retrieving one location at a time as shown in \cref{fig:example_old}. The search mode is highly limited for more complex use case, such as querying multiple relevant locations.
Spatial Exemplar Query (SEQ) offers a more flexible alternative by allowing users to query one or more nearby locations simultaneously with ease \cite{zhang2022example}. Users can specify search requirements either by general criteria similar to traditional search, or by a particular realistic example on the map. Then, candidate combinations of locations are identified by the similarity of individual location attributes as well as relative spatial distances. This approach aligns better with how users naturally explore new places, particularly in unfamiliar areas, rendering it useful for tasks such as travel planning. Previous studies find that SEQ is more efficient and user-friendly when performing complex multi-location tasks.

While SEQ offers more freedom for users' search queries, it is more challenging to precisely understand user needs and achieve favorable results. Due to the complexness of query criteria, the user input may be ambiguous or area-specific. For instance, urban railway transportation is referred to as \textit{`subway'} in the United States and as \textit{`MRT'} (Mass Rapid Transit) in Singapore. An American tourist searching for \textit{`subway'} in Singapore may be confused by the irrelevant results using a rigid system. Addressing these difficult cases usually necessitates manual annotation efforts in conventional search services.
Alternatively, from the user perspective, there is a common need of providing feedback on the current results, adjusting input criteria, and repeating the search for more satisfying locations. In the conventional design, each search query is performed independently, which does not guarantee improved quality in a sequence of queries.
Therefore, these practical requirements call for a more general and intelligent redesign of the spatial search system for a more user-friendly query experience.

\noindentparagraph{Integrating Language Models for Spatial Data.}
Large Language Models (LLMs) have emerged as a powerful tool for understanding sequential data and interacting with human users. Thanks to their proficiency in natural language and broad knowledge in various domains, these advanced deep learning models can easily adapt to a range of general tasks such as question answering and document retrieval. Furthermore, with appropriate configuration, LLMs are capable of scheduling and conducting a task sequence composed of subtasks based on user feedback in real time. In the realm of data management, LLMs have been utilized in diverse applications from parsing query languages to understanding database systems \cite{li2024b}.

The concept of incorporating LLMs for managing spatial data and assisting SEQ is presented in \cref{fig:example_seq}. Compared with other applications, this search mode encompasses several specific design goals:
(1) \textit{Data alignment} is crucial for assuring query quality. Current map data, including location coordinates, descriptions, and categories, are typically stored as structured data in database. While interacting with users in natural language, the LLM needs to extract necessary information from the conversation and query the database in a particular query language. Conversely, understanding the geospatial data structure is also essential for the LLM to reflect the provided results and revise search based on user feedback.
(2) \textit{Dialogue scheduling} is the novel ability brought by LLM for managing the overall search flow. Beyond a single search query, the LLM needs to recognize the current conversation status and user instruction for adjusting criteria, continuing, or rewinding the search. The unprecedented task requires tuning the LLM model for reliable performance.
(3) The utilization of LLM models needs to be budgeted to maintain a reasonable cost for the SEQ service.

\noindentparagraph{Our Approach.}
In this work, we present \name{}, which develops a new SEQ system by exploiting the power of LLM. Compared with conventional spatial search, \name{} highlights flexibility and generality throughout the search process, allowing for specifying the query, revising the criteria, and controlling the search process in natural language. It is also compatible with the standard SEQ search mode by directly specifying examples on the map.

To address the key design objectives in enhancing SEQ, \name{} employs a pair of specialized LLMs separately for spatial data processing and conversation tasks. The data alignment model is used to parse spatial data and call location searches in the backend. The dialogue scheduling model is responsible for deciding the function calls by interacting with users. We also develop sophisticated data adaptation schemes to facilitate LLMs in the task through data synthesis and model fine-tuning. Overall, the end-to-end design of \name{}, encompassing model learning, backend data processing, and frontend interface, provides a novel demonstration of applying LLMs for example-based spatial search in real-world applications.

\begin{figure}[!t]
    \centering
    \includegraphics[width=0.96\columnwidth]{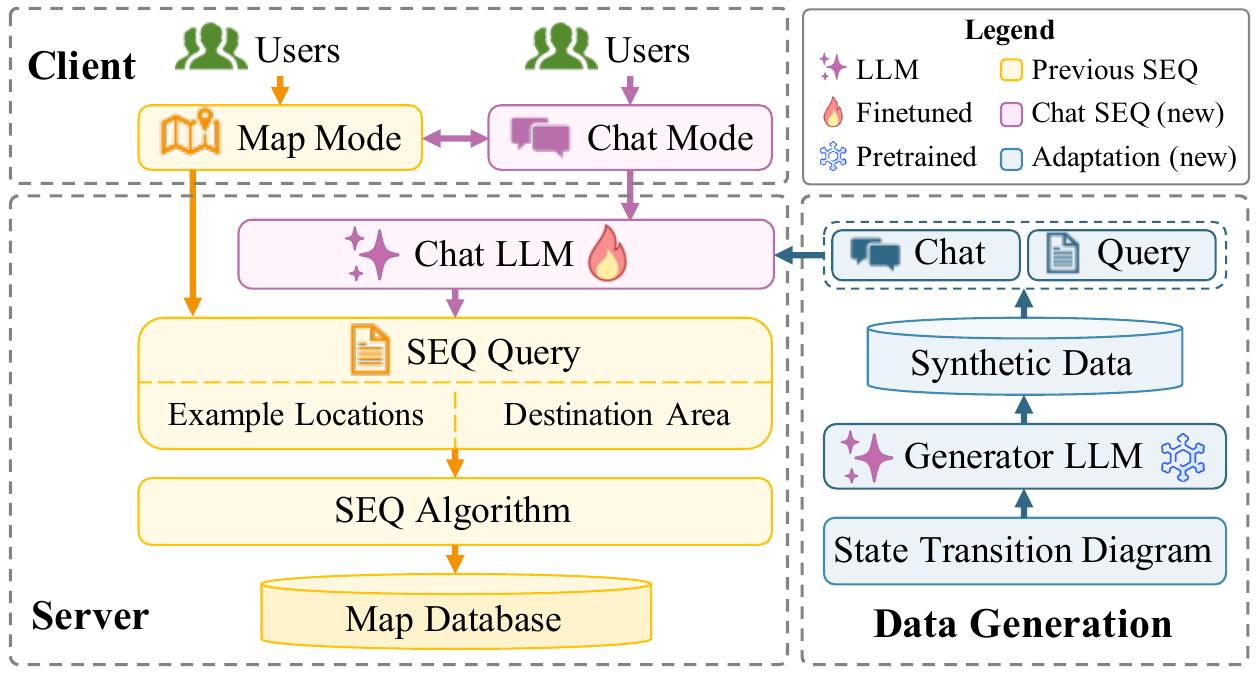}
    \caption{Overview of the \name{} architecture.}
    \label{fig:geobot-arch}
\vspace{-8pt}
\end{figure}

\section{\name{} System Overview}
\label{sec:system-overview}
\cref{fig:geobot-arch} presents the key components of \name{}. In a nutshell, \name{} employs a client-server system design. The server implemented in Gin Web Framework manages data processing and LLM communication tasks, while the React client captures inputs in the chat and map. The generator LLM for data synthesis is a pretrained model with frozen parameters, while the chat LLMs are smaller models with finetune adaptation to specific tasks.
In the following subsections, we respectively elaborate on the novel designs of \name{} on top of conventional spatial query systems, including model adaptation, spatial data processing, and user interactions.

\begin{figure*}[t]
    \centering
    \begin{subfigure}[b]{0.50\textwidth}
        \centering
        \includegraphics[height=54mm]{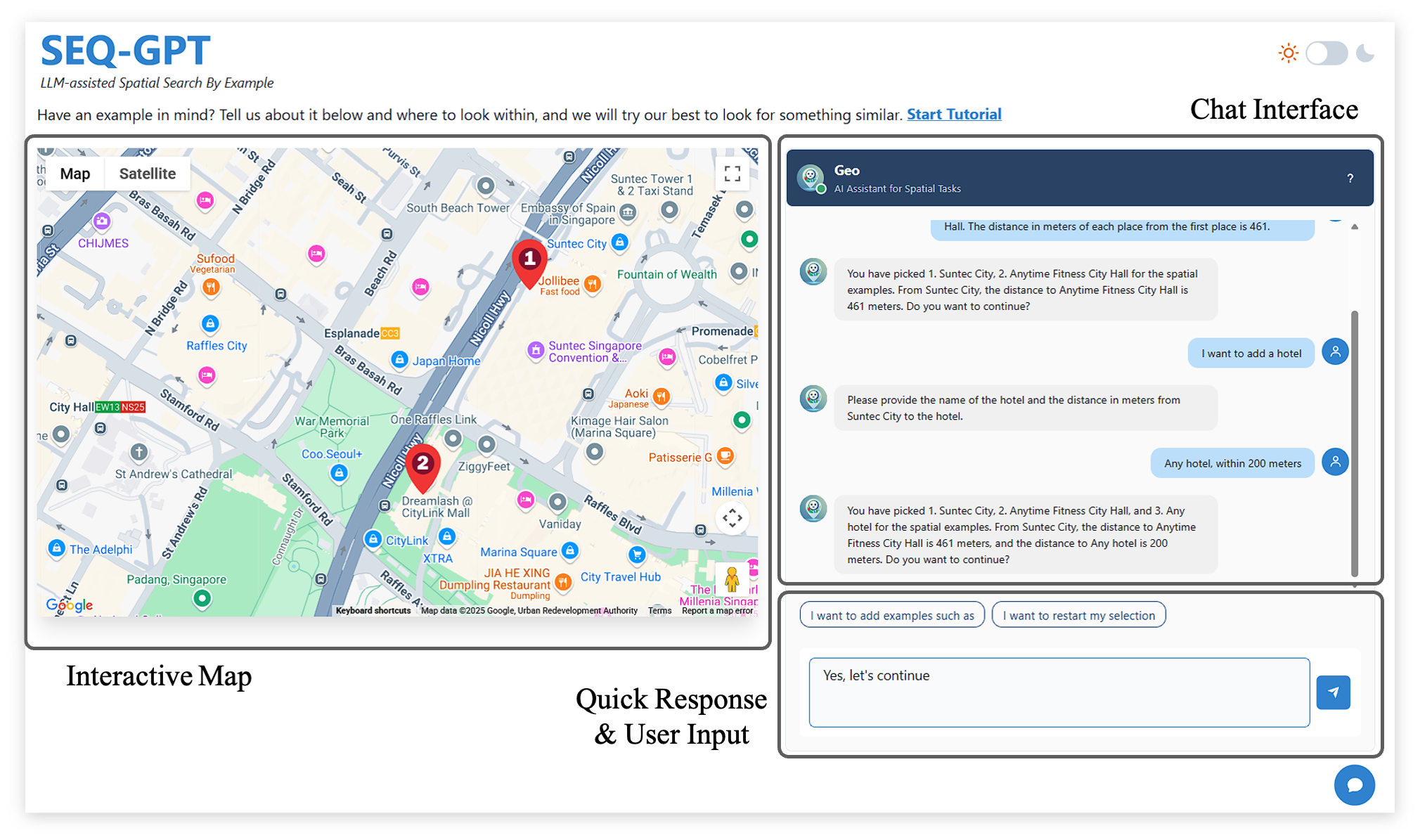}
        \caption{\name{} interface in Chat Mode.}
        \label{fig:demo-page}
    \end{subfigure}
    \hfill
    \begin{subfigure}[b]{0.48\textwidth}
        \centering
        \includegraphics[height=54mm]{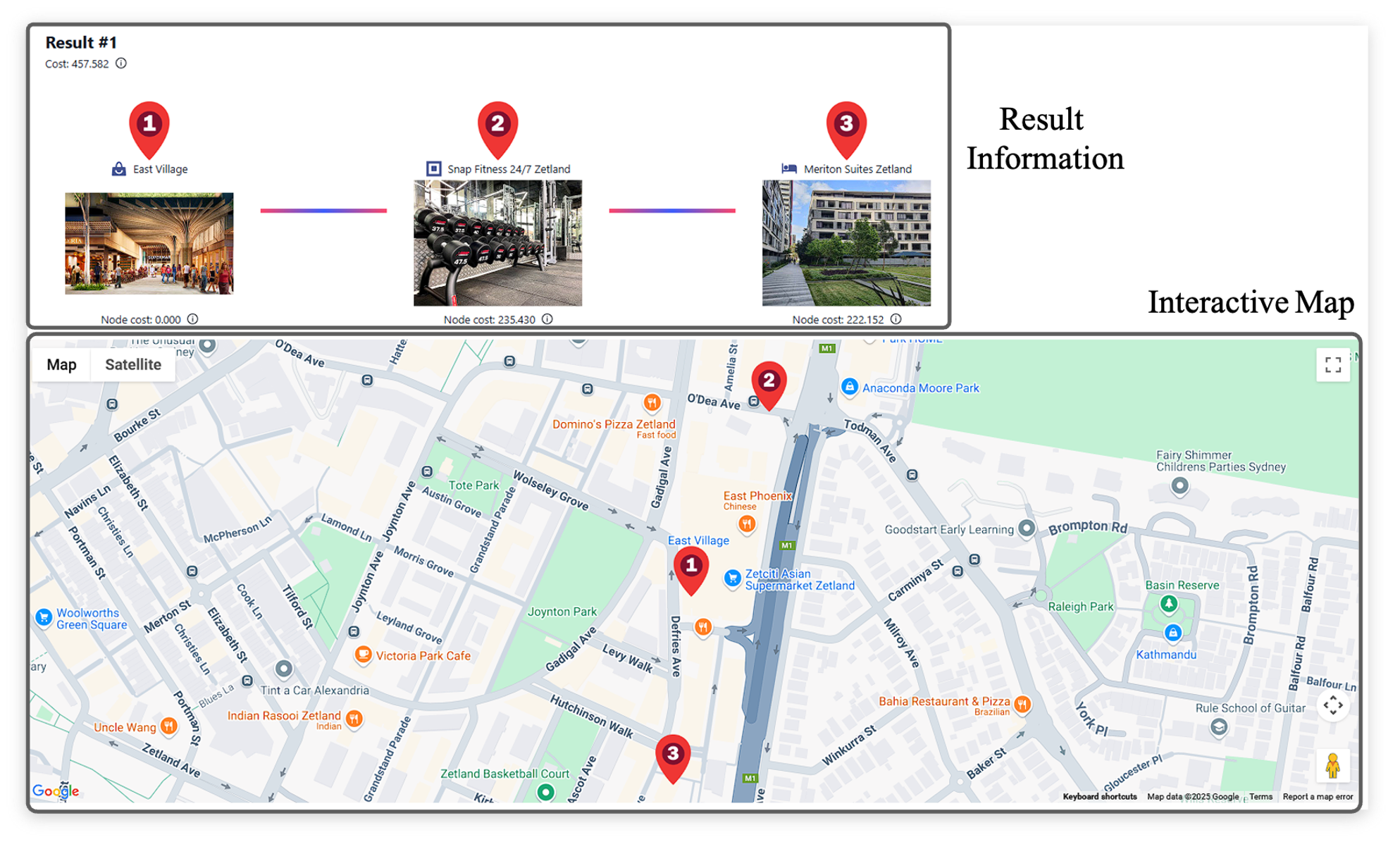}
        \caption{\name{} result display}
        \label{fig:demo-result}
    \end{subfigure}
\vspace{-2pt}
    \caption{Screenshots of (a) main user interface with an interactive map with marked location examples, and a chat area with conversation history and query tips; (b) search results in a list of locations and related information in the target area.}
    \label{fig:demo}
\vspace{-8pt}
\end{figure*}

\subsection{Spatial Query Processing}
\label{ssec:method-data}
\noindentparagraph{Synthetic Data Generation.}
Parsing the user dialogues into structured queries is the primary goal for LLM-assisted spatial data processing. However, the main difficulty lies in the lack of available dialogue datasets, which is essential for guiding the LLM to extract required outputs.
To this end, we employ the SynTOD data synthesizing pipeline \cite{samarinas2024simulating} for simulating realistic user-system interactions to generate dialogue data and corresponding formatted queries, which is used for LLM adaptation in \cref{ssec:method-llm}.

\begin{table}[!b]
    \caption{An example of properties, input prompt, and generated output of a conversation state.}
    \label{tab:state}
    \centering
    \vspace{-6pt}
\begin{tcolorbox}[
    enhanced,
    tabularx={p{11ex}|L},
    colback=Thistle!5!white,
    colframe=Thistle!75!black,
    boxrule=1pt,
    before upper app={\rowcolor{Thistle!15!white}}
]
    \textbf{State} & \texttt{spatial\_examples}
    \\\hline
    \textbf{Role} & \texttt{user}
    \\\hline
    \textbf{Prompt} & \small{You are a human user talking to a bot that helps with guiding users to pick some places. Provide exactly \{NUM\} spatial examples following the format $\cdots$}
    \\\hline
    \textbf{Text} & \textit{``I want to look for places like a gym and a station.''}
    \\\hline
    \textbf{Query} & [\textit{``gym''}, \textit{``station''}]
    \\\hline
    \textbf{Next State} & \texttt{select\_examples}
\end{tcolorbox}
\end{table}

\begin{figure}[!b]
\vspace{-8pt}
    \centering
    \includegraphics[width=\columnwidth]{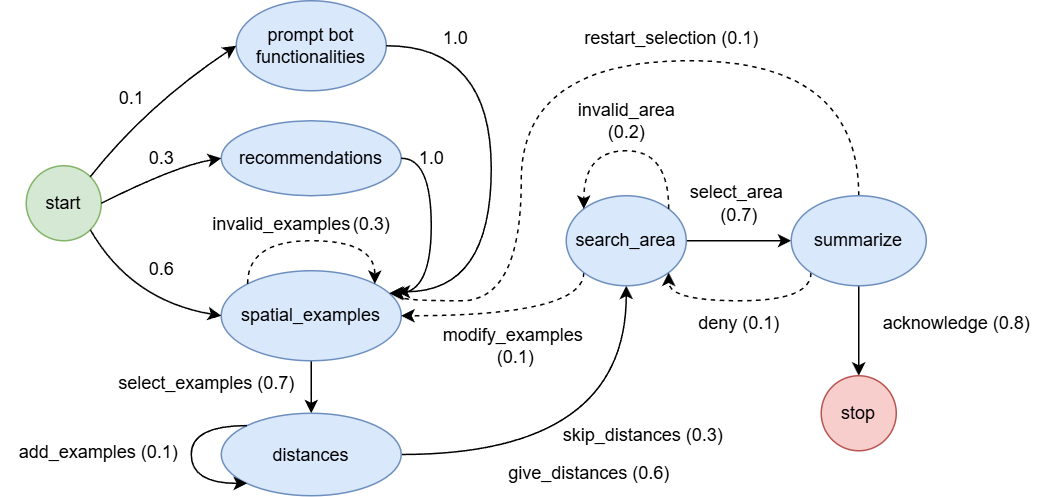}
    \caption{State transition graph illustrating possible user interactions with \name{} in data synthesis. Each node represents a chat state, and edges denote possible transitions between the states triggered by user interactions.}
    \label{fig:dialogue-state-diagram}
\end{figure}

We particularly develop the state transition model for our scenarios as shown in \cref{fig:dialogue-state-diagram}. The data generator performs constrained random walks on the transition graph representing different conversational states, such as introductions, request handling, and error resolution. The edge weight between states control indicates the transition probability.
Each state belongs to one of the three LLM roles, i.e., \texttt{system}, \texttt{user}, and \texttt{assistant}, where the \texttt{system} role provides high-level instructions for the the model behavior, and the other two respectively correspond to the roles in a real user-LLM dialogue. An LLM is used as generator for composing dialogue text according to the role and a tailored prompt for each state. An example state is shown in \cref{tab:state}, where the ``\texttt{text}'', ``\texttt{query}'', and ``\texttt{next state}'' fields are produced by the generator.

The process is repeated until reaching the \texttt{stop} state. All generated information in a walk are combined to form one training sample containing a pair of dialogue and query. The coherent and diverse dialogue data benefits the model in learning a meaningful extraction strategy in the context of SEQ, while ensuring the parsing results follow the required format.
In our implementation, we design 31 states in total, together with specialized prompts and transition patterns. The pretrained \textsf{GPT-4o} model is utilized as the generator to promote synthesis quality by its large model size.

\noindentparagraph{Proximity-based Search.}
The structured query for SEQ contains example location information such as category and relative distance, and the destination area. The SEQ algorithm \cite{luo2017seq} performs distance-based query using public available map APIs to acquire the combination of locations based on proximity. Since the number of candidate results are usually small in a given area, the search process can be efficiently finished.
Additionally, the system backend features a caching mechanism, which stores results of previous queries to speed up similar queries, since some result locations are highly determined and overlapped across repetitive searches.

\subsection{LLM Adaptation and Scheduling}
\label{ssec:method-llm}
\noindentparagraph{Model Finetuning and Evaluation.}
Since the correctness of query processing is crucial to the SEQ service, we specifically employ finetuning techniques to adapt the LLMs for \name{} deployment as chat models, thereby ensuring rigorous data format alignment. We enable the utilization of a variety of LLMs from different model families, including \textsf{GPT-4o-mini} and \textsf{LLaMA-3-8B-Instruct}. The corresponding adaptation strategies are Supervised Finetuning (SFT) and Parameter-efficient Finetuning (PEFT), respectively.

The finetuning dataset is synthesized by the process in \cref{ssec:method-data}, resulting a total of $2,000$ samples and are used for $3$ epochs of finetuning. Evaluation metrics including self-BLEU for diversity assessment, intent prediction accuracy, and relevance scoring are comprehensively utilized to select the suitable model.

\noindentparagraph{Model Coordination in Chat.}
After finetuning, the chat models are deployed for handling conversation. To deliver a precise and coherent SEQ experience, \name{} integrates multiple chat models, each finetuned for a subtask of the spatial search process, such as processing locations and defining search areas.
Coordination among the models are implicitly conducted by a signal-based protocol. At the end of each conversation state, the current chat model outputs a special token representing the predicted state, indicating whether to proceed to the next step or restoring to a previous state. This modular coordination allows for reacting to user inputs and scheduling the conversation accordingly.

\subsection{User Interface}
\noindentparagraph{Search Modes and Conversion.}
The frontend design of \name{} resembles some popular mapping services, aiming to facilitate familiar interaction from users. It offers two search modes: Map Mode is the classic style of SEQ search for giving examples by directly interacting with the map; Chat Mode is the novel interface uniquely available in \name{} powered by LLMs. Through this conversational interface, users can formulate spatial search queries naturally by providing examples or refining their preferences over multiple turns. Tailored responses from chat models are given for displaying the current conversation state and guiding user inputs.

The two modes between map and chat can be seamlessly converted, which further enhances the user experience by integrating the benefits of SEQ operations in both modes. Alternatively, users can choose different finetuned models as the chat model family to balance considerations of cost, scalability, and performance.

\noindentparagraph{Results Display.}
After SEQ computation in the backend server, result candidates are sent to the client and displayed on the map orienting the destination area. Each matched location in a result set will be presented in the designated sequence of the query, including the location name, image, category, and the similarity score, which is computed by the difference in geographical distance from the corresponding example location. A specific query usually leads to multiple result sets containing different locations in the destination area. These results are ranked and listed according to the similarity with the provided example sett. The user can click on each result to locate it in the map and expand detail information, or return to chat to give more search queries.

\section{Demonstration}
\label{sec:demo}
The demonstration examples showcases typical use cases of users performing SEQ search on the \name{} platform, featuring the unique experiences of spatial search by examples and the chat-based interaction. We display the core user interface and respective components in \cref{fig:demo}.

A user (\hlc[bubblegreen]{\textit{green}}) is planning a trip to an unfamiliar area and needs to find a hotel with convenient facilities. The user first employs the Map Mode as as \cref{fig:demo-page} and navigates to their current position to identify two examples based on familiar locations:  a shopping mall and a gym.
Then, the user clicks the mode conversion button to switch to Chat Mode. The input is automatically transferred to text and unfolds the following conversation with \name{} (\hlc[bubblegray]{\textit{gray}}):
\begin{rightbubbles}
    I want to search for places like 1. Suntec City and 2. Anytime Fitness City Hall. The distance in meters of each place from the first place is 416 meters.
\end{rightbubbles}
\begin{leftbubbles}
    You have picked 1. Suntec City and 2. Anytime Fitness City Hall. Do you want to continue and acknowledge the selection?
\end{leftbubbles}

Before continue, the user chooses to add a hotel as the third location. Instead of giving a specific example, an location in Chat Mode can also be queried by its category (hotel) and relative distance:
\begin{rightbubbles}
    I want to add a hotel
\end{rightbubbles}
\begin{leftbubbles}
    Please provide the name of the hotel you'd like to add, along with its distance in meters from the first place.
\end{leftbubbles}
\begin{rightbubbles}
    Any hotel within 200 meters
\end{rightbubbles}
\begin{leftbubbles}
    You have pick 1. Suntec City, 2. Anytime Fitness City Hall, 3. any hotel. From Suntec City, the distances are 461 meters to Anytime Fitness City Hall, and 200 meters to any hotel.
\end{leftbubbles}

By acknowledging the examples, \name{} proceeds to the next state to ask for target area. The position and search radius can be provided by the user through chatting in a similar fashion:
\begin{leftbubbles}
    Now I need you to provide a general search area to look within, like a neighborhood, city, region, or even a specific landmark.
\end{leftbubbles}
\begin{rightbubbles}
    In downtown Sydney
\end{rightbubbles}
\begin{leftbubbles}
    Your search area is valid, which is central Sydney, and the examples chosen are Suntec City, Anytime Fitness City Hall, and the hotel.

    I'm at hard work to find the best match!
\end{leftbubbles}

After specifying and confirming all search criteria, the SEQ search is performed and results are displayed as \cref{fig:demo-result}.


\section{Related Work}
\label{sec:related}
\noindentparagraph{Spatial Query.}
A common limitation of prevailing spatial query research is their reliance on spatial keyword queries and filtering~\cite{chen2013spatial}, which potentially restricts the search flexibility and applicability in various domains. In contrast, an array of works extends the query to a group of locations, especially for road networks, offering use case studies and dedicated query algorithms for SEQ \cite{luo2012disks,luo2014distributed}.
\textsc{Example Searcher} \cite{Yew2023example} demonstrates a system focusing on fast backend SEQ algorithms and criteria- or map-based search frontend, which is regarded as the standard SEQ schema in this study.

\noindentparagraph{Location-Based Service (LBS).}
Canonical studies in LBS primarily focus on spatial keyword queries, which are designed to optimize the retrieval of relevant results within a specified geographic area~\cite{lei2017topk,li2012direct}. This approach targets to provide users with location-specific information, enhancing the usability of LBS in various real-world scenarios. However, a limitation in current research is the focus on single-location searches per query, while the more flexible grouped search settings is largely underexplored.

\noindentparagraph{Large Language Model (LLM).}
Recent advances in LLM demonstrate the utilization of generative language models in various real-world applications, facilitating enhanced decision-making and problem-solving processes \cite{talebira2023multi}.
Moreover, there has been an innovative integration of LLM-driven technology in Geographic Information Systems (GIS) ~\cite{chen2023interact,li_autonomous_2023}.


\bibliographystyle{ACM-Reference-Format}
\bibliography{reference}

\end{document}